# A Sensitivity Analysis of Pathfinder

Keung-Chi Ng*
Bruce Abramson [†]
Department of Computer Science
University of Southern California
Los Angeles, CA 90089-0782

## Abstract

Knowledge elicitation is one of the major bottlenecks in expert system design. Systems based on Bayes nets require two types of information—network structure and parameters (or probabilities). Both must be elicited from the domain expert. In general, parameters have greater opacity than structure, and more time is spent in their refinement than in any other phase of elicitation. Thus, it is important to determine the point of diminishing returns, beyond which further refinements will promise little (if any) improvement. Sensitivity analyses address precisely this issue—the sensitivity of a model to the precision of its parameters. In this paper, we report the results of a sensitivity analysis of Pathfinder, a Bayes net-based system for diagnosing pathologies of the lymph system. This analysis is intended to shed some light on the relative importance of structure and parameters to system performance, as well as the sensitivity of a system based on a Bayes net to noise in its assessed parameters.

## 1 Introduction

Over the past ten years or so, expert systems—large, knowledge-intensive artificial intelligence (AI) programs designed to aid decision makers and diagnosticians—have become increasingly important at many medical, corporate, industrial, military, educational, and other institutions. The success of an expert system, like that of all software, is dependent on two factors, its ability to generate accurate (and relevant) information and its ability to communicate that information effectively. Communication between the system and its users is generally embodied in the *user interface;* its data is stored in the *knowledge base* and manipulated by the *inference engine.* There is a great deal of literature about interface design [4] [13], and much has been said about the propriety of competing inference techniques [5] [22] [27] [33], uncertainty management mechanisms [23], and algorithms for propagating belief [15] [21] [24]. Until quite recently, general discussions about knowledge bases have been conspicuously sparse, and even then focused almost entirely on knowledge acquisition. There are, however, other interesting aspects of a knowledge base. This paper investigates an expert system's *sensitivity* to precision in the knowledge that it encodes.

Knowledge acquisition is frequently the bottleneck in system design. Few experts can immediately translate information that they acquired through years of training and experience into machine-understandable form. In general, iterative refinements are necessary to obtain a relatively accurate knowledge base. The sensitivity issue, then, is an important one. When does the refinement process hit a point of diminishing returns? How "good" must the encoded information to be considered "good enough"? An expert system, like many other activities, may hit a *flat maximum,* or a point beyond which minor changes in input variables are highly unlikely to produce major changes in the system's output [31]. If a flat maximum is reached, there is little point in attempting to refine the expert's assessments. Sensitivity analyses address the potential merits of these further refinements—if a knowledge base can be shown to be insensitive to the introduction of noise, it is unlikely to be helped by the removal of noise. Statements about system sensitivity, of course, can never be made in a vacuum; different domains will show different sensitivities. This paper presents an analysis of Pathfinder, a medical system for lymph node pathology [11] [12].

*Supported in part by the National Library of Medicine under grant R01LM04529.

[†]Supported in part by the NSF under grant IRI-8910173.

## 2 Bayes Nets and Pathfinder

In the past, few sensitivity analyses have been performed on expert systems, primarily because few systems have been amenable to such analyses. Many "classic" expert system knowledge bases were configured as a set of *production rules*, a formalism that was originally popularized because of its proximity to formal logic and procedural knowledge. Just as a logical proposition is either true or false, however, the connection between a rule's precedent and its antecedent is either valid or it is not—"minor" changes are almost impossible to define. Most designers of rule-based systems found pure rules to be representationally inadequate, and thus augmented them with some form of numeric correlation between precedent and antecedent. Pathfinder belongs to what may be termed the next generation of expert systems. Rather than augmenting rules with numbers, Pathfinder's primary representation scheme is numeric—the system's underlying format emerges from the relationships among these numbers.

Pathfinder's underlying model is known as a *Bayes net* [24]. A Bayes net, (and its close relative, the influence diagram [6] [16] [25] [26]), is a graphical representation of a domain in which variables are modeled as nodes, relationships are modeled by arcs, and arc weights define the strength of that conditionality. In the lymphnode pathology domain of Pathfinder, for example, the nodes represent diseases and symptoms, and the arcs indicate dependence between diseases and symptoms, or among symptoms. An arc from disease A to symptom B indicates that the probability with which a patient will exhibit symptom B is dependent on whether or not he has disease A. The weight on the arc specifies that conditional probability. Pathfinder's disease nodes combine to specify a mutually exclusive and exhaustive set of *hypotheses;* any patient being diagnosed by the system is assumed to have exactly one of the candidate diseases. The task of the diagnostician is to collect evidence that discriminates among them. Thus, at any point in time, Pathfinder's partial diagnosis may be represented as a probability distribution over all the hypotheses. A diagnosis is complete when the probability assigned to one of the diseases equals (or approaches) 1.0.

Every component of a Bayes net must be elicited by the system designer from an expert; variables (represented as nodes) and their possible values, relationships (represented as arcs), and degrees of influence (represented as probabilities) must all be specified. In addition, *prior* probabilities must be specified across the hypotheses, corresponding to the probability that each disease will be present, given no patient specific information. In many of these areas, however, the expert's initial responses are frequently incorrect. If any expert could sit down and construct an accurate Bayes net, system design would be relatively simple. Unfortunately, few experts are used to thinking of their domains in terms amenable to Bayes net modeling.[1] A trained analyst is necessary to help guide them through the process, to pose questions that the experts may have overlooked themselves, and to make sure that the resultant models are internally consistent [17]. Some aspects of modeling are more difficult than others. Devising a list of variables is frequently quite simple. Thus, nodes can generally be created rather quickly. The range of possible values taken on by these variables is a bit trickier, but relatively uncontroversial. The existence of a connection between two variables is rarely difficult to make, and most experts can easily provide the information necessary to draw arcs. The determination of appropriate conditional probability distributions to serve as arc weights, however, is quite difficult and time consuming. Increased efficiency in probability assessments could greatly reduce the time needed to design a system.

---

[1] Discussions with David Heckerman and Dr. Bharat Nathwani indicate that Pathfinder's expert, Dr. Nathwani, was quite comfortable specifying probabilities. Psychological evidence [19], however, indicate that this skill is rare.

The idea that ballpark probabilities may be as powerful as carefully refined assessments is not as radical as it sounds. An analysis of MYCIN [28], a rule-based expert system that used certainty factors, indicated that the certainty factors associated with the rules were not very important to the system's performance— MYCIN's diagnoses were, in general, identical with the certainty factor engine on or off [3, pages 217–219]. In the clinical psychology literature on bootstrapping, Dawes described a series of tests in which he compared judgments made directly by experts to those generated by linear model of the experts. In addition to the standard, regression weighted models, Dawes included some improper linear models in which the weights were either random or uniform. The result was that all linear models outperformed raw expert judgment [7]. Abramson's experiments on chess material-advantage functions yielded similar results [1] [2]; in a 28,000 game tournament among expert-designed, regression-learned, uniform, and randomly generated weights, no single set was able to demonstrate superiority over the others.

## 3 Sensitivity Analysis

The goals of a sensitivity analysis are (i) to gain insight into the nature of a problem, (ii) to find a simple and elegant structure that does justice to the problem, and (iii) to check the correctness of the numbers and the need for precision in refining them [31, page 387]. Although this characterization of a sensitivity analysis should be familiar to decision analysts, it may appear somewhat unusual to people from other fields. In most decision problems, once the numbers have reached a certain degree of precision, further refinement on these numbers has little effect on the decisions. Whether similar observations are true for diagnostic problems, (i.e., once the prior and conditional probabilities have reached certain quality, further improvement on these probabilities has little effect on its diagnoses), such as Pathfinder's, is the subject of this paper.

The information stored in a Bayes net can be divided into two components:

1. Structure: nodes and arcs.

2. Parameters: prior and conditional probabilities.

Network structure plays an obvious role in system performance. The full extent of structure's importance, however, remains to be established. Both Dawes' improper linear models [7] and Abramson's tournaments [2] indicate that, at least at times, structure is almost the sole determining factor of strength. This study was devised to examine the role of parameters to Pathfinder's performance.

Experiments were run on a body of 60 "classic" cases in which the diagnosis was known. Since a network's parameters include prior and conditional probabilities, both sets of probabilities had to be varied. The experiments reported in the next section used two sets of prior probabilities (those specified by the experts and a uniform distribution across the hypotheses) and three types of conditionals:

1. The original values, exactly as assessed by experts.

2. Randomly generated probabilities. This class of parameters includes probabilities distributed both uniformly and normally.

3. The values assessed by experts plus randomly generated noise, using both uniformly and normally distributed noise functions.

Each body of tests served a different purpose; the original knowledge base define a standard against which others may be judged, the random parameters addressed the relative importance of structure and parameters, and the random noise addressed the issue of sensitivity. The use of two different sets of priors addressed the effect of priors on system performance.

## 4 Report of Results

We found that system performance degraded so significantly with randomly generated probabilities that the resulting system had negligible discriminating power. These results were observed regardless of the distribution function used for generating the conditional probabilities or the selection of priors. These findings led us to conclude that parameters are crucial to a Bayes net (or at least to Pathfinder's Bayes net) and that experts are needed to provide the parameters.

Having shown that parameters play an important role in system performance, our next task was to assess the quality of Pathfinder's parameters with respect to their sensitivity to noise. Some minor changes have been made to some conditional probabilities used in Pathfinder during system evaluation, to cater for mis-diagnoses on several test cases [9], indicating that refining parameters can lead to improvement in performance. Our analysis was intended to assess all of Pathfinder's parameters.

Our experiments studied variations in both prior and conditional probabilities. Priors were fixed either at the expert-assessed set or at a uniform set. Conditionals were varied by augmenting the expert's assessment with randomly generated noise. The resultant conditional probabilities are then renormalized. The random noise functions followed uniform or normal distributions with $\mu = 0$ and several values of $\sigma$. For each noise function, five parameter sets were created. Sixty cases were run on each network, for a total of 300 data points per noise function. A total of 7 noise generating schemes were used, including uniformly generated noise (uniform noise), normally generated noise (normal noise) with standard deviations ($\sigma$) of 0.005, 0.01, 0.025, 0.05, 0.1, 0.25, all with a mean ($\mu$) of 0 (to ensure that any probability has equal chance of being increased or decreased). A summary of the test results with expert priors is shown in Table 1. In the table, the percentage of correct diagnoses, (i.e., the number of cases in which the known diagnosis was assigned the highest probability divided by the total number of cases), is intended to provide a measure of the diagnostic power. The average confidence, (average difference in posterior probabilities between the two diseases with the highest posterior probabilities on the differential diagnosis for all the cases run), with respect to the correct diagnosis[2] and incorrect diagnosis, provides a measure of the discriminating power of the leading disease (disease with the highest posterior probability on the differential diagnosis) from the other diseases on the differential diagnosis. It should be pointed out that although the percentage of correct diagnosis is more important than average confidence in system performance, average confidence is also useful in gauging system performance. Systems scoring perfectly (100%) in the correct diagnosis column with 0 average confidence (e.g., all diseases have the same posterior probability with respect to all the test cases) could be as useless a system as one with no correct diagnoses and absolute average confidence (1.0). Also shown in Table 1 (in the column headed "Percentage Better"), is the percentage of cases in which the noisy network assigned the correct diagnosis with a higher probability than did the original network.

Table 1 indicates that the original knowledge base had the highest score in both percentage of correct diagnoses and average confidence; augmentation with uniform noise produced the lowest scores on both items. Adding normal noise to the original knowledge base produced a system with scores that lie between these extremes, with better results for systems with smaller standard deviations (or less noise). Further-

---

[2]The disease with the highest posterior probability on the different diagnosis provided by the system is the same as the known diagnosis for a test case.

208| | Percentage Correct | Average Confidence | | Percentage Better |
|---|---|---|---|---|
| | | Correct (# cases) | Incorrect (# cases) | |
| Original knowledge base | 98.3% | 0.7910 (59) | 0.8247 (1) | — |
| Normal noise, $\sigma=0.005$ | 90.0% | 0.7329 (270) | 0.2511 (30) | 22.0% |
| Normal noise, $\sigma=0.01$ | 87.6% | 0.6963 (263) | 0.3685 (37) | 21.6% |
| Normal noise, $\sigma=0.025$ | 78.0% | 0.6803 (234) | 0.2692 (66) | 16.3% |
| Normal noise, $\sigma=0.05$ | 70.7% | 0.6997 (212) | 0.2541 (88) | 15.7% |
| Normal noise, $\sigma=0.1$ | 61.0% | 0.6212 (183) | 0.3262 (117) | 9.3% |
| Normal noise, $\sigma=0.25$ | 36.7% | 0.5614 (110) | 0.2988 (190) | 6.0% |
| Uniform noise | 32.7% | 0.5417 (98) | 0.3142 (202) | 5.0% |

Table 1: Summary of results of the sensitivity analysis of Pathfinder. In this set, expert-assessed priors were used for all networks. A variety of noise functions were added to the expert's conditional probabilities.

more, the chance of producing a better diagnosis than the original knowledge base is higher for knowledge bases with less noise than those with more noise. All these observations suggest that Pathfinder's knowledge base is of high quality and performs well. Table 2 summarizes the results of networks with equal priors. The results are similar to those of Table 1.

The results summarized in Tables 1 and 2 strengthen the evaluations of Pathfinder reported in [12]. In addition to corroborating the system's diagnostic power, our experiments confirmed both the expert's probability assessments and the effort placed on refining them; even marginal noise degraded the system, and further refinements would have been unlikely to improve it. Of course, this does not necessarily mean that Pathfinder—or any comparable system—avoids flat maxima. Standard decision analytic sensitivity analyses tend to be of the single-fault variety; only one parameter is varied at a time. The typical definition of a flat maximum, then, is a situation where minor perturbations in *any* of the system's inputs is unlikely to have a major impact on its output. Our studies showed that even minor changes in *all* of Pathfinder's parameters can have a substantial impact on performance. Our decision to vary all of the system's parameters simultaneously reflects one of the changes that AI imposes on DA models: increased size. Pathfinder's underlying network—like that of all Bayes net knowledge bases—is much larger than the models typically used in DA. This size increase motivated Pathfinder's designers to introduce *similarity networks* as a workable way of specifying Bayes nets [10]; standard constructions were viewed as unwieldy. In a similar manner, we felt that there were too many variables for individual analysis. Aggregate (simultaneous) analysis was sufficient to address the issue of under-refinement vs. over-refinement (or, as our results showed, proper refinement).

In a certain sense, then, our studies were not completely fair. In Pathfinder, as in most systems, variables are not of equal importance, and a system's sensitivity to noise is unlikely to be uniform across its variables. One of the items that we observed during our analysis was that some diseases were less susceptible to misdiagnosis by the addition of noise. The diagnosis of a disease like hairy cell leukemia, for example, is based entirely on the presence of hairy cells. Noisy probabilities associated with that one key symptom have a significant negative impact on the system's ability to detect the disease. More complicated diagnoses, in which many features play a role, are less susceptible to the vagaries of a single noisy datum. This observation is not surprising; it occurs in many test settings. In a complex collection of logic gates, for example, simple functions calculated by passing through a single gate will be completely unreliable if that gate is faulty. More complex functions, however, may be robust enough to be recovered. In the same way, simple and obvious diseases will be misdiagnosed if noise is introduced in the wrong place. Complicated diagnoses are more robust. Since these are precisely the data in which an expert's assessments are least likely to err, analyses like the one described above may be biased against the system.

|  | Percentage Correct | Average Confidence | | Percentage Better |
|---|---|---|---|---|
|  |  | Correct (# cases) | Incorrect (# cases) |  |
| Original knowledge base | 88.3% | 0.8262 (53) | 0.1569 (7) | — |
| Normal noise, $\sigma=0.005$ | 84.7% | 0.7900 (254) | 0.1279 (46) | 24.3% |
| Normal noise, $\sigma=0.01$ | 83.0% | 0.7688 (249) | 0.1444 (51) | 17.7% |
| Normal noise, $\sigma=0.025$ | 80.0% | 0.7019 (240) | 0.2052 (60) | 18.3% |
| Normal noise, $\sigma=0.05$ | 73.7% | 0.6784 (221) | 0.1831 (79) | 13.7% |
| Normal noise, $\sigma=0.1$ | 62.3% | 0.6458 (187) | 0.3019 (113) | 10.0% |
| Normal noise, $\sigma=0.25$ | 42.3% | 0.4914 (127) | 0.2516 (173) | 7.3% |
| Uniform noise | 36.7% | 0.4347 (110) | 0.2927 (190) | 6.0% |

Table 2: Summary of results of the sensitivity analysis of Pathfinder. In this set, uniform priors were used for all networks. A variety of noise functions were added to the expert's conditional probabilities.

## 5 Conclusion

In this paper, we have shown that parameters play an important role in Pathfinder's performance; of these parameters, conditional probabilities are more important than priors. Our analysis of Pathfinder confirmed the diagnostic prowess claimed by its designers. In addition, they indicated a well-performed elicitation; the assessed probabilities were accurate enough to produce near-optimal performance. Although these results, in and of themselves, may not appear earth-shattering, they do highlight an important point: outsiders (i.e., people other than the system's designers) were able to investigate and experimentally validate a knowledge engineering exercise. This type of experimentation is rare in AI and almost unheard of in knowledge engineering; it was possible, in large part, because of the transparency of the Bayes net formalism.

Verifiable, reproducible, and controlled experimentation is an important part of science, and it is one of the areas in which AI has been traditionally weak [1]. The recent wave of work on Bayes nets, however, has suggested several different types of experiments: comparisons of different uncertainty formalisms [8], competitions between Bayes nets and rule bases [14] [20] [30] [32], and several different approaches to (and motivations for) sensitivity analyses [18] [29]. For the most part, these studies address the behavior of a system; although they are all system-specific, they should have some general implications to the way in which we approach system design. Our results, for example, suggest that future system designers consider their underlying model's sensitivity to noisy parameters before expending time and effort on parameter refinement. We believe that stronger results should be possible, and we hope to see many of the experimental techniques of behavioral psychology modified to investigate knowledge-based systems. Our sensitivity analyses represent what we hope is one step towards the development of reproducible controlled experiments for AI systems.